\pdfoutput=1

\documentclass[11pt]{article}

\usepackage[review]{EMNLP2023}

\usepackage{times}
\usepackage{latexsym}

\usepackage[T1]{fontenc}

\usepackage[utf8]{inputenc}

\usepackage{microtype}

\usepackage{inconsolata}
\usepackage{graphicx}

\title{The Radiation Oncology NLP Database}

\author{First Author \\
  Affiliation / Address line 1 \\
  Affiliation / Address line 2 \\
  Affiliation / Address line 3 \\
  \texttt{email@domain} \\\And
  Second Author \\
  Affiliation / Address line 1 \\
  Affiliation / Address line 2 \\
  Affiliation / Address line 3 \\
  \texttt{email@domain} \\}

\begin{document}
\maketitle
\begin{abstract}
We present the Radiation Oncology NLP Database (ROND), the first dedicated Natural Language Processing (NLP) dataset for radiation oncology, an important medical specialty that has received limited attention from the NLP community in the past. With the advent of Artificial General Intelligence (AGI), there is an increasing need for specialized datasets and benchmarks to facilitate research and development. ROND is specifically designed to address this gap in the domain of radiation oncology, a field that offers many opportunities for NLP exploration. It encompasses various NLP tasks including Logic Reasoning, Text Classification, Named Entity Recognition (NER), Question Answering (QA), Text Summarization, and Patient-Clinician Conversations, each with a distinct focus on radiation oncology concepts and application cases. In addition, we have developed an instruction-tuning dataset consisting of over 20k instruction pairs (based on ROND) and trained a large language model, CancerChat. This serves to demonstrate the potential of instruction-tuning large language models within a highly-specialized medical domain. The evaluation results in this study could serve as baseline results for future research. ROND aims to stimulate advancements in radiation oncology and clinical NLP by offering a platform for testing and improving algorithms and models in a domain-specific context. The ROND dataset is a joint effort of multiple U.S. health institutions. The data is available at \url{https://github.com/zl-liu/Radiation-Oncology-NLP-Database}.
\end{abstract}

\section{Introduction}

Radiation oncology is a critical medical specialty that employs high-energy radiation to treat and manage cancer and other diseases \cite{bernier2004radiation,unkelbach2018robust}. Indeed, like many medical domains, there is much potential to integrate natural language processing (NLP) into radiotherapy research and practice \cite{bitterman2021clinical,rezayi2022clinicalradiobert}. However, there is limited development and evaluation of NLP models in this domain due to the lack of dedicated datasets \cite{rezayi2022clinicalradiobert}. In response to this need, we present the Radiation Oncology NLP Database (ROND).

\begin{figure*}[t]
 	\centering
 \includegraphics[width=\textwidth]{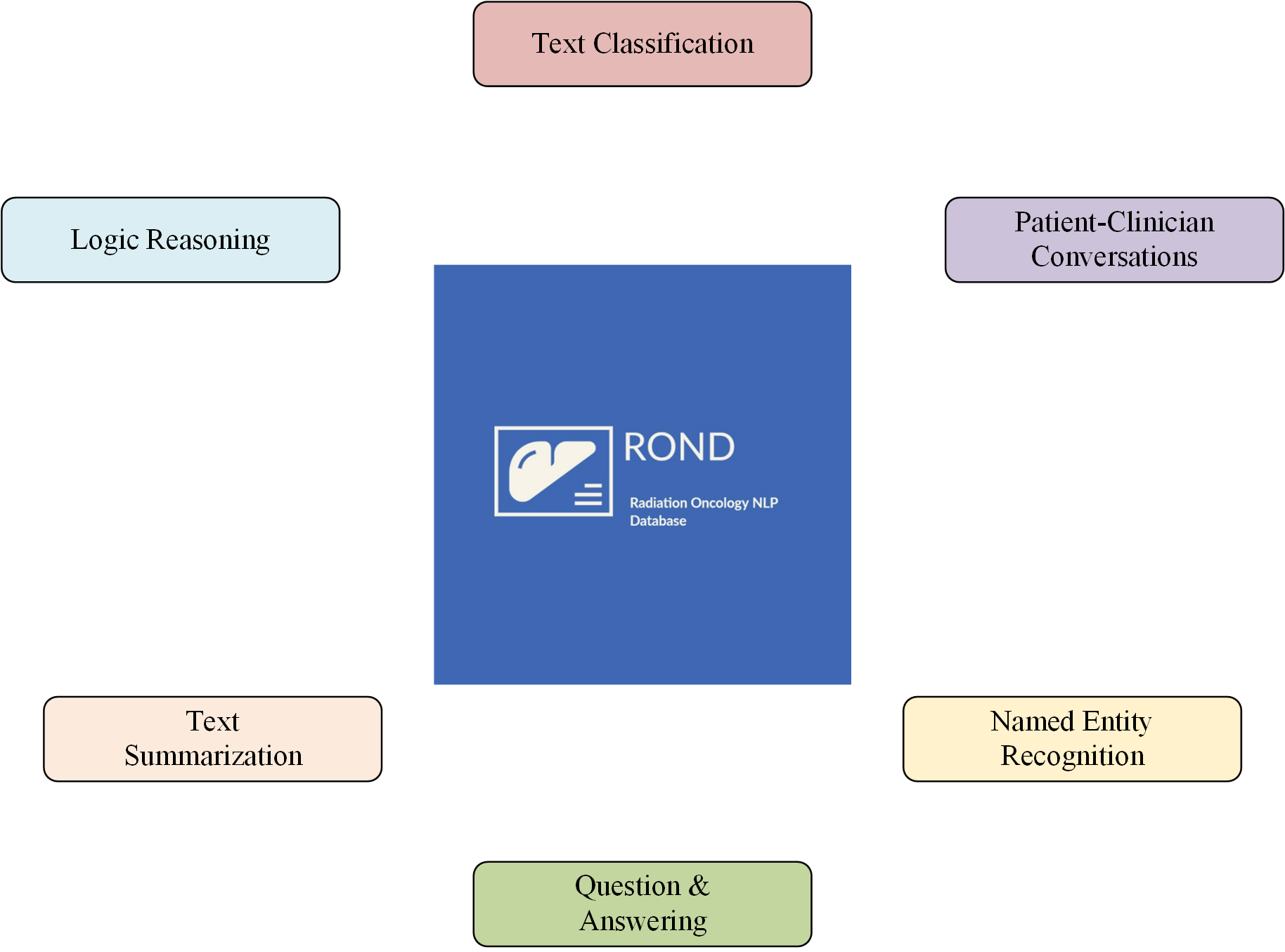} 
    \caption{Overview of the Radiation Oncology NLP Database.}
    \label{fig:main}
\end{figure*}

ROND is the world's first NLP dataset specifically created for radiation oncology. It aims to provide a comprehensive platform for researchers to develop, test, and improve NLP models and methods within this domain. This dataset covers a wide spectrum of NLP tasks, including Logic Reasoning, Clinical Text Classification, Named Entity Recognition (NER), Question Answering (QA), and Text Summarization. Each of these tasks is centered around distinct aspects of radiation oncology, offering researchers a rich and varied dataset for exploration and model training. In addition, ROND contains a Patient-Clinician conversation dataset, which provides valuable insights into patient interactions, symptom descriptions, and treatment discussions, enhancing our understanding and modeling of complex medical dialogues. Figure \ref{fig:main} presents an overview of ROND.

The unique structure of ROND facilitates the development of models capable of reasoning logically about complex radiation oncology concepts, classifying domain-specific text data, recognizing and categorizing specialized entities, accurately answering radiation oncology-related questions, and summarizing lengthy documents and research papers in the field. We aim to establish a benchmark for future studies that stimulates innovation in radiation oncology research and ultimately improves patient care through the power of NLP.

We believe this database is of particular importance in the age of Artificial General Intelligence (AGI) \cite{bubeck2023sparks,zhao2023brain,liu2023summary}. Successful large language models (LLM) such as ChatGPT, GPT-4, LLAMA \cite{touvron2023llama} and PaLM \cite{chowdhery2022palm} are trained on vast amounts of public domain data. Some LLMs such as Med-Palm 2 \cite{singhal2022large} are trained on both public biomedical data sources and private hospital (e.g., through the Google-Mayo Clinic partnership) data, and consequently are highly capable of processing medical text \cite{singhal2022large}. However, there is no existing dataset that specifically supports NLP in radiation oncology. The ROND dataset complements recent LLM advancements and offers a platform to better integrate LLMs into healthcare.

\section{The Radiation Oncology NLP Database}

\subsection{Logic Reasoning}
The Logic Reasoning subset of the Radiation Oncology NLP Database (ROND) presents questions designed to assess the logical reasoning capabilities of NLP models within the context of radiation oncology. The questions are geared towards the understanding of fundamental concepts and principles in radiation oncology, such as the properties of radioactive elements, atomic structure, electron orbits, X-ray emission, penumbra effects, and interaction of different particles with matter. We manually created and annotated 100 logic reasoning questions for this dataset. 

Each question in this subset is structured as a yes/no question, designed to elicit a binary response. The questions range from basic atomic structure, such as "Does an atom consist of a positively charged nucleus surrounded by a cloud of negatively charged electrons?" to more specific queries about X-ray production and penumbra effect, such as "In X-ray production, does the efficiency of x-ray production depend on the size of the target?" or "Is physical penumbra influenced by geometric penumbra, beam energy, and the lateral transport of electrons in the tissues?"

This dataset provides an avenue to evaluate the ability of NLP models to apply logical reasoning within the domain-specific context of radiation oncology, emphasizing both the understanding of fundamental radiation oncology concepts and the ability to apply this knowledge to specific scenarios.

\begin{figure*}[t]
 	\centering
 \includegraphics[width=\textwidth]{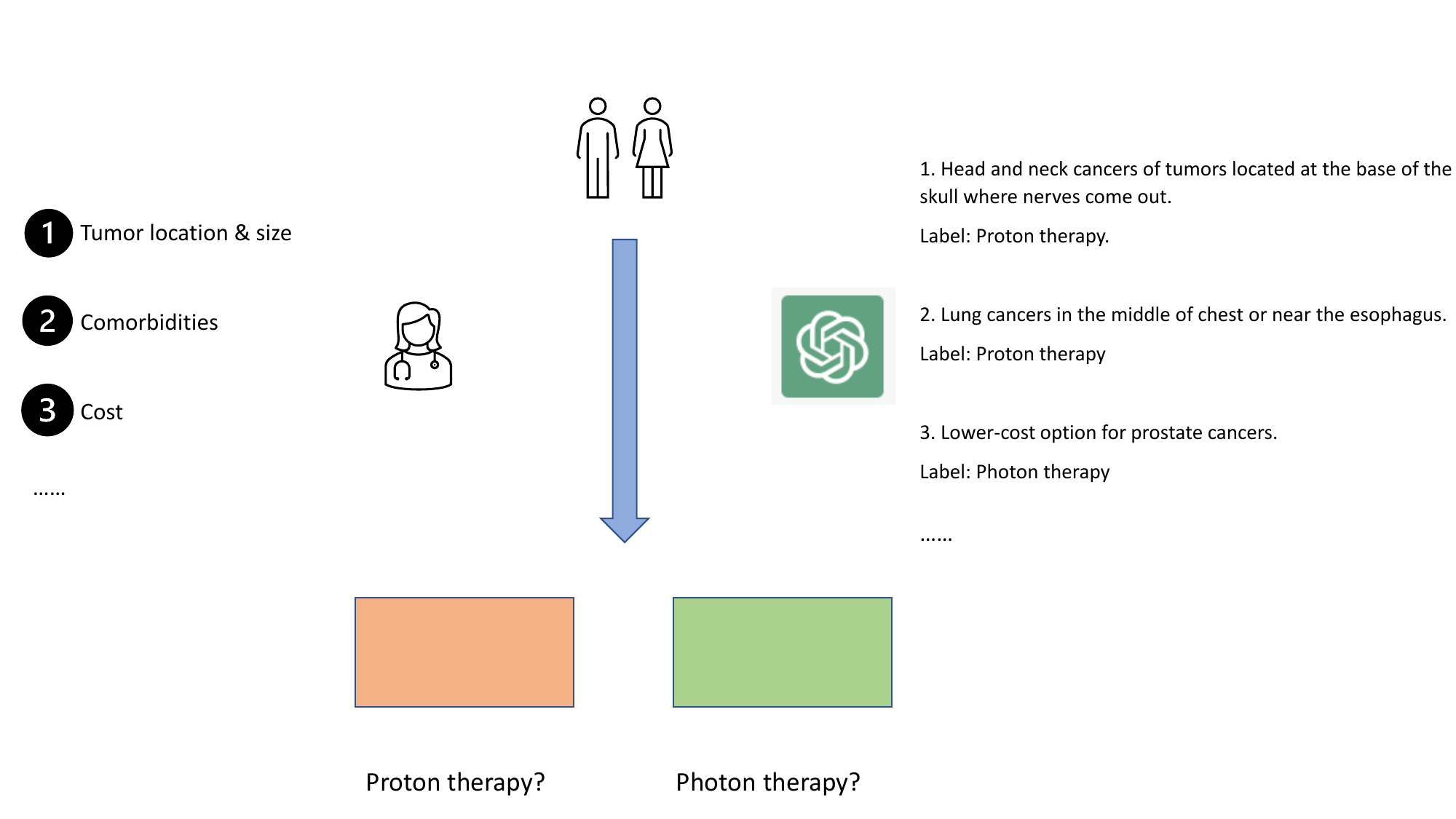} 
    \caption{Illustration of the Clinical Text Classification dataset.}
    \label{fig:clinical_significance_text_classification}
\end{figure*}

\subsection{Clinical Text Classification}
The Clinical Text Classification subset of ROND is designed to test the capability of NLP models in categorizing text inputs related to radiation oncology into predefined labels. This specific task focuses on determining the appropriate type of therapy ("Proton therapy" or "Photon therapy") based on descriptions of different cancer scenarios. We manually created and annotated 100 cases for this dataset. Please see Figure \ref{fig:clinical_significance_text_classification} for an illustration of this dataset. 

The dataset presents a variety of clinical scenarios and characteristics of cancers, such as the location and sensitivity of the tumor, cost considerations, patient demographics, and potential risks. These descriptions are then categorized into two major classes: "Proton therapy" and "Photon therapy".

Examples include categorizing "Head and neck cancers of tumors located at the base of the skull where nerves come out" and "Cancers in children" under the label 'Proton therapy'. On the other hand, "Lower-cost option for prostate cancers" and "Better protection of skin" are classified under 'Photon therapy'.

We aim to facilitate the training and evaluation of NLP models capable of accurately classifying radiation oncology cases into relevant treatment venues, thereby potentially aiding decision-making processes in clinical settings.

\begin{figure*}[t]
 	\centering
 \includegraphics[width=\textwidth]{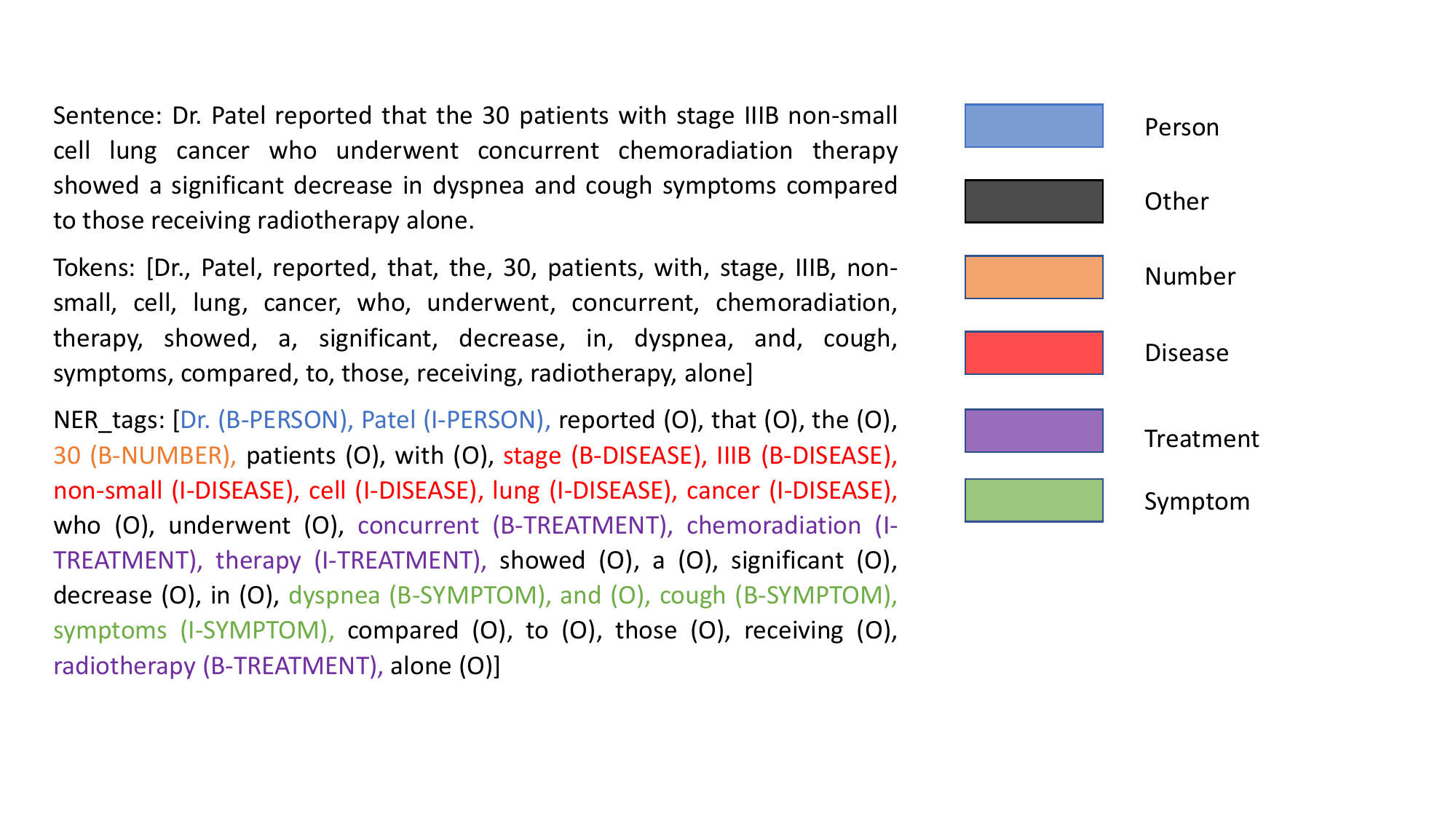} 
    \caption{A sample of the NER dataset.}
    \label{fig:ner_colored_sample}
\end{figure*}

\subsection{Named Entity Recognition (NER)}
The Named Entity Recognition (NER) subset of ROND is designed to annotate entities in the text that pertain to the field of radiation oncology. This task is crucial for understanding specific details within the text, such as identifying the names of doctors and patients (PERSON), types of diseases (DISEASE), types of treatment (TREATMENT), anatomical structures (ANATOMY), numeric values (NUMBER), symptoms (SYMPTOM), and outcomes (OUTCOME). We asked GPT-4 to generate sample sentences, and manually reviewed and annotated 20 sets of NER samples that are factually correct. While not all sentences necessarily contain all seven NER tags, these are the maximum possible tags that any sentence from our sample sets might include. Figure \ref{fig:ner_colored_sample} presents a sample from the NER dataset. 

For instance, in the sentence "Dr. Jenkins, a radiation oncologist, treated patient Sarah Williams for breast cancer, utilizing intensity-modulated radiation therapy (IMRT) with a total dose of 50 Gy in 25 fractions", the model is expected to identify "Dr. Jenkins" as a PERSON, "radiation" and "intensity-modulated radiation therapy" as a TREATMENT, "breast" as an ANATOMY, "50" and "25" as NUMBERs, and "skin irritation" as a SYMPTOM. 

In the context of another sentence, "In a study led by Dr. Jackson, 60 patients with glioblastoma were treated using hypofractionated radiation therapy, administering 40 Gy in 15 fractions", the model should detect "Dr. Jackson" as a PERSON, "glioblastoma" as a SYMPTOM, "hypofractionated radiation therapy" as a TREATMENT, and "40" and "15" as NUMBERs.

The objective of this subset is to evaluate a model's ability to identify these entities in radiation oncology text, which is fundamental for structured information extraction and other downstream tasks such as de-identification of sensitive patient information (e.g., names and addresses) \cite{liu2023deid}.

\subsection{Text Summarization}
The Text Summarization dataset within the Radiation Oncology NLP Database offers a unique set of resources for the exploration of text summarization methods in a highly specialized medical context. This dataset comprises a variety of research abstracts from arXiv that are categorized under the Medical Physics class. First, we collected all such papers published since 2022. We then programmatically extracted the abstracts from these papers and asked GPT-4 to produce summaries. Finally, we manually selected 200 paper abstracts that are correctly and meaningfully summarized by GPT-4 to form this dataset. 

Each record within the dataset includes key bibliographic information such as the title of the research, the authors involved, the submission date, the arXiv identifier, the DOI, and the BibTeX entry. The categorization information is also provided in the form of classifications.

The core components of each record are the abstract and its corresponding summary. The abstracts provide a brief, yet comprehensive overview of the research conducted, including its objectives, methodology, results, and conclusions. Correspondingly, the summaries distill the critical elements of these abstracts into a concise form, designed to swiftly provide the reader with the key takeaways of the study.

These pairs of abstracts and their summaries constitute a valuable resource for supervised learning tasks. They can facilitate the development and fine-tuning of models focused on abstract summarization within the domain of radiation oncology and medical physics, a critical area in the broader field of cancer treatment.

\begin{figure*}[t]
 	\centering
 \includegraphics[width=\textwidth]{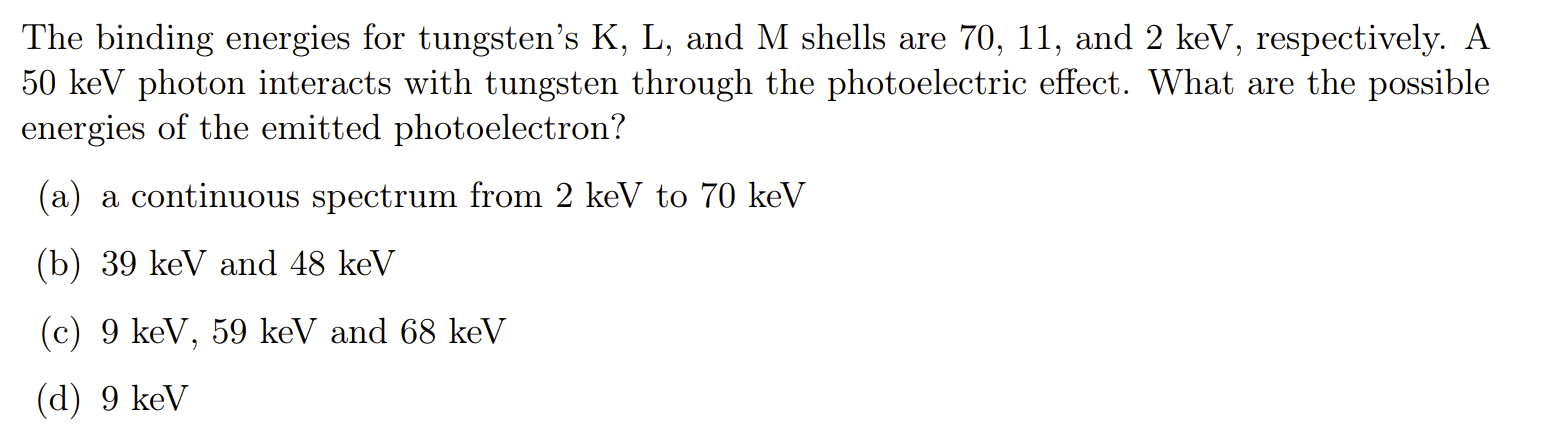} 
    \caption{A sample multiple-choice question from the QA dataset.}
    \label{fig:qa_sample}
\end{figure*}

\subsection{Question and Answering (QA)}
The QA subset of the Radiation Oncology NLP Database stands as a rigorous and comprehensive collection of multiple-choice questions encompassing a vast array of topics within the field of radiation physics. We in-house designed 100 questions comparable to those in the RAPHEX exam \cite{hendee2007abr}, which is a radiation oncology physics test-preparation exam for medical physicists and radiation oncologists. We created these questions from scratch, as we are not legally permitted to reproduce or redistribute materials from the RAPHEX exam. The questions cover eight categories: "math-based questions", "basic physics", "radiation measurements", "treatment planning", "imaging modalities and applications", "brachytherapy", "advanced treatment planning", and "safety, QA, and protection". Figure \ref{fig:qa_sample} contains an example from the QA dataset. 

Each question in this dataset demands a deep understanding of core radiation physics principles. Topics range from particle acceleration and atomic mass structures to photon interactions, x-ray spectra, radiation attenuation, and the functionalities of a linear accelerator. The multiple-choice format adds a layer of complexity to the challenge, necessitating that AI models not only comprehend the underlying physics concepts but also discern the most accurate answer from a selection of closely related options.

This dataset is designed to mimic the stringent conditions of academic assessments in the field. This design aids in creating a realistic test of an AI model's abilities in knowledge comprehension, reasoning, and mathematical computation under conditions mirroring those in a professional or educational context.

\begin{figure*}[t]
 	\centering
 \includegraphics[width=\textwidth]{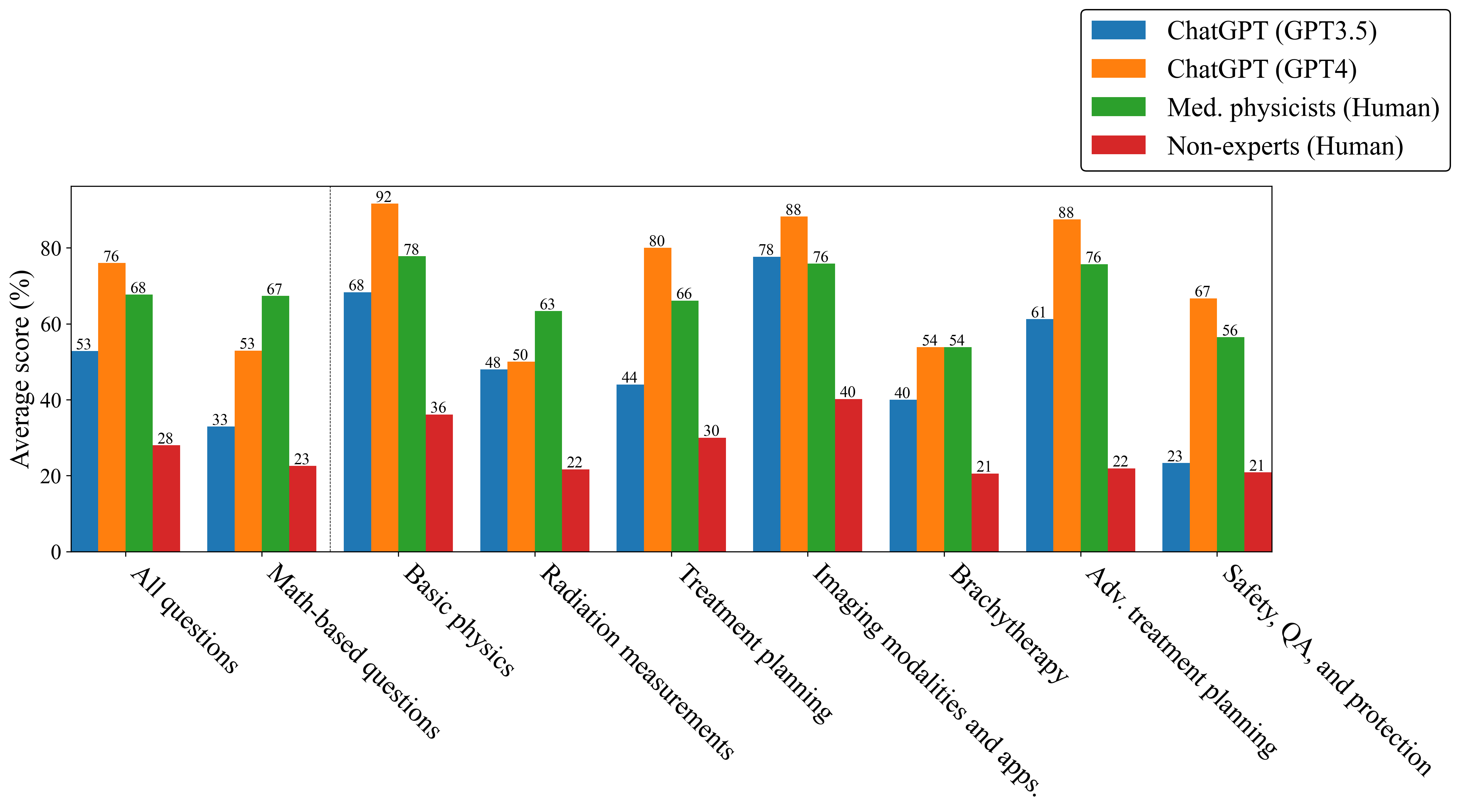} 
    \caption{A detailed analysis of LLMs' performance on the medical physics board exam (RAPHEX) level questions.}
    \label{fig:score_results}
\end{figure*}

\section{Clinical significance}
\noindent \textbf{Logic Reasoning:} The development of models capable of logic reasoning in radiation oncology holds the promise of aiding professionals in complex decision-making processes. Such models can enhance the comprehension of intricate clinical scenarios, support decision-making, and optimize individualized treatment planning. This would be particularly useful in areas such as proton therapy where detailed reasoning often guides the selection between different treatment strategies \cite{liu2012robust,unkelbach2018robust,schild2014proton}.

\noindent \textbf{Clinical Text Classification:} By training models to classify clinical cases, we are opening avenues to more personalized and efficient patient care. For instance, classifying patient cases into categories such as eligibility for photon versus proton therapy based on patient information and clinical notes can expedite decision making and improve treatment outcomes \cite{bitterman2021clinical,taylor2023prioritizing}.

\noindent \textbf{Named Entity Recognition (NER):} NER tasks, which involve identifying and classifying key information in text, provide a structured way to extract critical data points from unstructured clinical notes. This functionality is crucial in radiation oncology where specific entities, such as tumor types, anatomical locations, or dosimetric parameters, are paramount for the creation of optimal treatment plans \cite{unkelbach2018robust,schild2014proton,liu2018small}.

\noindent \textbf{Text Summarization:} The ability to extract the most salient information from large volumes of text is valuable in any field \cite{el2021automatic}, but in radiation oncology, it can directly contribute to improved patient care. For example, summarizing the key findings of the latest research in radiation oncology could help clinicians stay updated with current knowledge without having to go through lengthy papers, enabling them to swiftly apply these findings in their practice. In addition, text summarization can produce succinct descriptions from lengthy clinical reports and notes \cite{feblowitz2011summarization,cai2021chestxraybert,liu2023deid}, which can significantly save time and facilitate clinical communication. 

\noindent \textbf{Question and Answering (QA):} QA systems in the domain of radiation oncology could revolutionize the way practitioners access and analyze relevant information. Being able to ask specific questions and receive accurate answers quickly, whether in terms of patient history, or intricate radiobiological effects \cite{wang2018molecular,omer2021radiobiological}, would immensely improve the efficiency of the oncology practice, saving practitioners time and possibly enhancing patient outcomes. The QA dataset in this database delineates the knowledge needed for effective QA in this highly specialized domain.

\begin{table}[t]
\centering
\begin{tabular}{lll}
\hline
Bard  & ChatGPT & GPT-4 \\ \hline
0.600 & 0.456   & 0.656 \\ \hline
\end{tabular}

	\caption{Accuracy of state-of-the-art LLMs on the Logic Reasoning 
dataset.}
	\label{table1}
\end{table}

\begin{table}[t]
\centering
\begin{tabular}{lll}
\hline
Bard  & ChatGPT & GPT-4 \\ \hline
0.770 & 0.740   & 0.840 \\ \hline
\end{tabular}
	\caption{Accuracy of state-of-the-art LLMs on the Text Classification
dataset.}
	\label{table2}
\end{table}

\begin{table}[t]
\centering
\begin{tabular}{lll}
\hline
Bard  & ChatGPT & GPT-4\\ \hline
0.667 & 0.642   & 0.785 \\ \hline
\end{tabular}
	\caption{Accuracy of state-of-the-art LLMs on the NER
dataset.}
	\label{table3}
\end{table}

\begin{table}[t]
\centering
\begin{tabular}{lll}
\hline
Bard  & ChatGPT & GPT-4 \\ \hline
0.139 & 0.270   & 0.317 \\ \hline
\end{tabular}
	\caption{BLEU4 scores of state-of-the-art LLMs on the Text Summarization
dataset.}
	\label{table4}
\end{table}

\begin{table}[t]
\centering
\begin{tabular}{lll}
\hline
Bard  & ChatGPT & GPT-4\\ \hline
0.410 & 0.530   & 0.760 \\ \hline
\end{tabular}
	\caption{Accuracy of state-of-the-art LLMs on the QA
dataset.}
	\label{table5}
\end{table}

\begin{table}[t]
\centering
\begin{tabular}{ll}
\hline
ChatGPT & CancerChat \\ \hline
26      & 24  \\ \hline
\end{tabular}
	\caption{Preference evaluation between ChatGPT and CancerChat.}
	\label{table6}
\end{table}

\section{A Conversational Instruction Tuning Dataset based on ROND}

\begin{figure*}[t]
 	\centering
 \includegraphics[width=\textwidth]{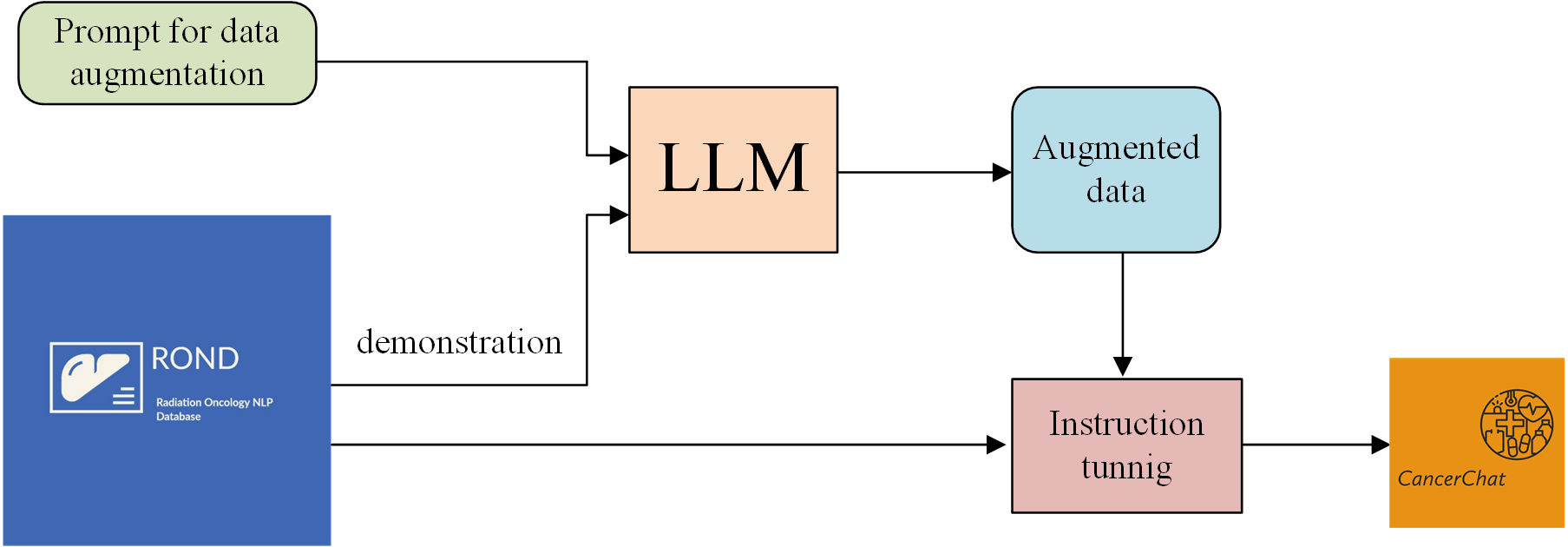} 
    \caption{Data augmentation and instruction tunning.}
    \label{fig:data_aug}
\end{figure*}

To facilitate future development of radiation oncology focused language models, we employed a data generation method to create synthetic data based on expert annotated data from the six key components within ROND: Logic Reasoning, Clinical Text Classification, Named Entity Recognition (NER), Text Summarization, Question and Answering (QA), and Conversational data.

This synthetic data generation process was carried out to build a large dataset suitable for "instruction tuning". Instruction tuning is an effective approach aimed at enhancing language models' ability to comprehend and follow natural language inputs based on training on pairs of instruction-input-outputs \cite{wei2021finetuned,peng2023instruction}. This strategy facilitates multi-task learning and enhances generalization for unseen tasks \cite{ouyang2022training}.

\begin{figure*}[t]
 	\centering
 \includegraphics[width=\textwidth]{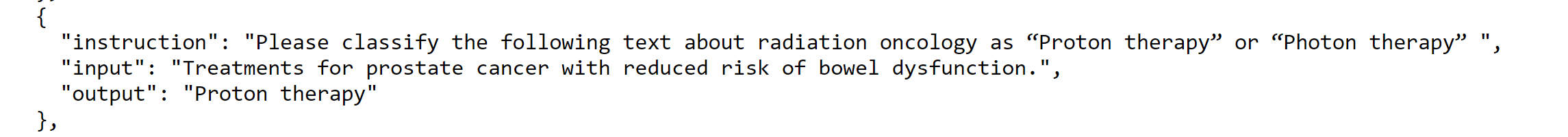} 
    \caption{An example from the instruction tuning dataset.}
    \label{fig:sample-instruction}
\end{figure*}

We processed the diverse data formats within ROND into a unified instruction tuning structure, where each data entry consists of three components: instruction, input, and output (please refer to Figure \ref{fig:sample-instruction} for an example). This uniform structure streamlines the training and tuning process, contributing to more efficient learning and inference.

For the data augmentation/synthetic data generation, we utilized the APIs of ChatGPT and GPT-4. Specifically, we employed the ChatGPT API to generate synthetic samples for the Conversational data, Text Summarization, and Named Entity Recognition. These tasks require somewhat less domain-specific understanding of radiation oncology, making ChatGPT a fitting choice. On the other hand, for tasks such as Logic Reasoning, Clinical Text Classification, and Question Answering that require more depth of understanding and reasoning ability, we used the GPT-4 API. This decision leverages the strength of each AI model, ensuring optimal synthetic data generation across different tasks within the ROND database.

We scaled ROND from a few hundred data samples to 20,160 instruction tuning pairs. The majority of these samples, over 17,000, are conversational, simulating interactions between patients and healthcare providers. This is supplemented with 500 samples each for Logic Reasoning and Clinical Text Classification, and 1,000 samples each for Named Entity Recognition and Text Summarization. This ensures a comprehensive dataset that reflects the multifaceted nature of language processing in the context of radiation oncology, thereby providing a robust base for developing and refining future NLP models in this domain. Furthermore, the emphasis on conversational data equips models trained on this dataset provides the potential to create efficient chatbots specialized in radiation oncology.

\subsection{CancerChat}
To evaluate the effectiveness and utility of the instruction tuning dataset, we trained a demo local LLM, CancerChat, based on Falcon-7B \cite{falcon40b} (a model that recently tops the Open LLM Leaderboard \cite{huggingfaceOpenLeaderboard}). Figure \ref{fig:data_aug} illustrates the process of generating the instructiom-tuning dataset and the training of CancerChat. 

The training was conducted on a server with 1 A100 80GB GPU. We utilize LoRA (Low Rank Adaptation) \cite{hu2021lora} since LoRA weights facilitate model sharing and deployment. Our training parameters were: a batch size of 128, a fixed learning rate at 3e-4, a lora\_r (the rank of the low-rank factorization) set to 8, a lora\_alpha (the scaling factor for the rank) set to 16, and a dropout rate of 0.05 to mitigate overfitting. 

To preliminarily assess the performance of CancerChat, we conducted a blind comparison using 50 queries (see Table \ref{table6}). A medical physicist was asked to evaluate and express their preference for the responses generated by ChatGPT and CancerChat. The comparison yielded close results, with 26 of the 50 responses more favorable for ChatGPT, while the remaining 24 preferred CancerChat.

These preliminary results underscore the potential of CancerChat, which, despite being a smaller local model specifically tailored for the domain of radiation oncology, demonstrated competitive performance when compared to ChatGPT. This suggests the feasibility and promise of developing domain-specific language models leveraging our instruction-tuning dataset, offering a pathway for future advancements in NLP for medical specialties.

\section{Benchmarking Large Language Models}
We evaluate three state-of-the-art language models, namely Bard (powered by Google PaLM 2 \cite{anil2023palm}), ChatGPT \cite{openaiIntroducingChatGPT}, and GPT-4 \cite{openai2023gpt}, against the proposed radiation oncology NLP database and obtained insightful results. 

On the Logic Reasoning dataset, GPT-4 achieved the highest accuracy of 0.656, followed by Bard with an accuracy of 0.600, while ChatGPT scored the lowest with an accuracy of 0.456 (Table \ref{table1}). 

On the Clinical Text Classification dataset, GPT-4 again outperformed the other two models with an accuracy of 0.840. However, the performance gap was narrower with Bard and ChatGPT registering accuracies of 0.770 and 0.740, respectively (Table \ref{table2}). Similar trends were observed in the Named Entity Recognition (NER) dataset, where GPT-4 led with an accuracy of 0.758. Bard scored 0.667, and ChatGPT trailed with an accuracy of 0.646 (Table \ref{table3}).

For text summarization (evaluated using the BLEU score \cite{papineni2002bleu}), GPT-4 recorded the highest score of 0.317, with ChatGPT performing markedly better than Bard, scoring 0.270 compared to Bard's 0.139 (Table \ref{table4}). Finally, on the Question Answering (QA) dataset, GPT-4 once again demonstrated superior performance with an accuracy of 0.760. ChatGPT achieved an accuracy of 0.530, while Bard scored the lowest with 0.410 (Table \ref{table5}). We also conducted a detailed analysis on eight categories of the QA dataset. Generally, GPT-4 outperformed both trained medical physicists and non-experts who participated in this study. The medical physicists averaged an accuracy of 0.76, while the non-experts averaged 0.28. 

These results underscore the effectiveness of GPT-4 across various NLP tasks. However, while GPT-4 demonstrated superior performance across the evaluated NLP tasks in ROND, we hold these results as a baseline for future research. As state-of-the-art as these models may be, we envision the development of more refined models specifically tailored to the domain of radiation oncology. These specialized models will likely exhibit improved performance and understanding of the nuances of this unique field and deliver significant potential for advancing radiation oncology and clinical NLP.

\section*{Limitations}
The primary limitation of this study is inherent to its novelty. ROND represents the first dedicated NLP dataset in the field of radiation oncology. As such, it is anticipated that there will be inherent challenges and unforeseen issues associated with this dataset that will only become apparent when the database is utilized practically by the \textbf{community}. The efficacy and practical utility of language models trained or evaluated on this dataset will need to be validated through real-world application and extensive feedback from \textbf{clinicians}. This iterative process of application, feedback, and refinement is essential to not only identify potential problems but also to improve the robustness and applicability of the models derived from this dataset.

\section*{Ethics Statement}
The authors of this study conducted the research in strict adherence to ethical guidelines and there is no involvement of Protected Health Information (PHI). We acknowledge that while this study has the potential to greatly improve radiation oncology NLP, it is necessary to consider the need for responsible use and development of AI. Any potential bias or errors within the dataset can have significant implications on model training and subsequent applications. Thus, we recommend rigorous validation, including feedback from clinical practitioners, before deploying models trained on this dataset in a practical setting. Future research should also be conducted with the broader social and ethical implications in mind, always prioritizing the safety and best interests of patients.

\bibliography{emnlp2023}
\bibliographystyle{acl_natbib}

\appendix



\end{document}